\documentclass[11pt]{article}
\usepackage{acl-hlt2011}

\usepackage{amsmath} 
\usepackage{latexsym} 
\usepackage{epsf} 
\usepackage[all]{xy}  
\usepackage{lscape} 
\usepackage{verbatim}    
\usepackage{times}  
\usepackage{color}
\usepackage{epsfig}
\usepackage{multirow}
\usepackage{url}

\title{Experimenting with Transitive Verbs in a DisCoCat}

\author{Edward Grefenstette \\
  University of Oxford\\ 
  Department of Computer Science \\
  Wolfson Building, Parks Road \\
  Oxford OX1 3QD, UK \\
  {\tt edward.grefenstette@cs.ox.ac.uk} \\\And
  Mehrnoosh Sadrzadeh \\
  University of Oxford\\ 
  Department of Computer Science \\
  Wolfson Building, Parks Road \\
  Oxford OX1 3QD, UK \\
  {\tt mehrs@cs.ox.ac.uk} \\}

\date{}
\begin{document}

\maketitle

%\ \vspace{-4cm}

\begin{abstract}
Formal and distributional semantic models offer complementary benefits in modeling meaning.
The categorical compositional distributional  model of meaning  of \newcite{Coeckeetal} (abbreviated to DisCoCat in the title) combines aspects of both to provide a general framework in which meanings of words, obtained distributionally, are composed using methods from the logical setting to form sentence meaning. Concrete consequences of this general abstract setting and applications to empirical data are under active study \cite{Grefenetal,GrefenetalEMNLP}. In this paper, we extend this study by examining transitive verbs, represented as matrices in a DisCoCat. We discuss three ways of constructing such matrices, and evaluate each method in a disambiguation task developed by \newcite{GrefenetalEMNLP}. 
\end{abstract}

\section{Background}

The categorical distributional compositional model of meaning of \newcite{Coeckeetal} combines the modularity of formal semantic models with the empirical nature of vector space models of lexical semantics. The meaning of a sentence is defined to be the application of its grammatical structure---represented in a type-logical model---to the kronecker product of the meanings of its words, as computed in a distributional model. The concrete and experimental consequences of this setting, and other models that aim to bring together the logical and distributional approaches,  are active topics in current natural language semantics research, e.g.~see~\cite{Grefenetal,GrefenetalEMNLP,Daoud,Baroni,Guevara,Lapata}.

In this paper, we focus on our recent concrete DisCoCat model~\cite{GrefenetalEMNLP} and in particular on nouns composed with transitive verbs. Whereby the meaning of a transitive sentence `sub tverb obj' is obtained by taking the component-wise multiplication of the matrix of the verb with the kronecker product of the vectors of subject and object:
\begin{equation}\label{eq}
 \overrightarrow{\text{sub tverb obj}} = \underline{\text{tverb}} \odot (\overrightarrow{\text{sub}} \otimes \overrightarrow{\text{obj}})
\end{equation}
In most logical models,  transitive verbs are modeled as relations; in the categorical model the relational nature of such verbs gets manifested in their matrix representation: if subject and object are each $r$-dimensional row vectors in some space $N$, the  verb will be a $r\times r$  matrix in the space $N \otimes N$. There are different ways of learning the weights of this matrix. In~\cite{GrefenetalEMNLP}, we developed and implemented  one such method on the  data from the British National Corpus. The matrix of each verb was constructed by taking the sum of the kronecker products of all of the subject/object pairs linked to that verb in the corpus. We refer to this method as the \emph{indirect method}. This is because the weight $c_{ij}$ is obtained from the weights of the subject and object vectors (computed via co-occurrence with bases $\overrightarrow{n}_i$ and $\overrightarrow{n}_j$ respectively), rather than directly from the context of the verb itself, as would be the case in lexical distributional models. This construction method was evaluated against an extension of \newcite{Lapata}'s disambiguation task from intransitive to transitive sentences.  We showed and discussed how and why our method, which is moreover scalable and respects the grammatical structure of the sentence,  resulted in better results than other known models of  semantic vector composition. 

As a motivation for the current paper, note that there are  at least two different factors at work in Equation (\ref{eq}): one is the matrix of the verb, denoted by $\underline{\text{tverb}}$, and the other is the kronecker product of subject and object vectors $\overrightarrow{\text{sub}} \otimes \overrightarrow{\text{obj}}$. Our model's mathematical formulation of composition  prohibits us from changing the latter kronecker product, but the `content' of the verb matrices  can be built through different procedures. 

In recent work  we used a standard lexical distributional model for nouns and engineered our verbs to have a more sophisticated structure because of the higher dimensional space they occupy.  In particular,  we argued that the resulting matrix of the verb should represent `the extent according to which the verb has related the properties of  subjects to the properties of its objects', developed a general procedure to build such matrices, then studied their empirical consequences. One question remained open: what would be the consequence of starting from the standard  lexical vector of the verb, then encoding it into the higher dimensional space using different (possibly ad-hoc but nonetheless interesting)  mathematically inspired methods. 

%Starting with the direct context vector of the  verb, what we would be computing is 
%\[
%(*) \quad \overrightarrow{\text{tverb}} \odot (\overrightarrow{\text{sub}} \otimes \overrightarrow{\text{obj}})
%\] 
%Whatever the answer, it will for once and for all crown a winner in the battle between vectors and matrices for transitive verbs of the categorical model?
In a nutshell, the lexical vector of the verb is denoted by  $\overrightarrow{\text{tverb}}$ and similar to vectors of subject and object, it is an $r$-dimensional row vector. Since  the kronecker product of subject and object $(\overrightarrow{\text{sub}} \otimes \overrightarrow{\text{obj}})$ is $r\times r$,  in order to make $\overrightarrow{\text{tverb}}$ applicable in Equation~\ref{eq}, i.e.~to be able to substitute it for $\underline{\text{tverb}}$,   we need to encode it  into a   $r\times r$ matrix in the $N \otimes N$ space. In what follows, we investigate the empirical consequences of three different encodings methods.

\section{From Vectors to Matrices}

In this section, we discuss three  different ways of encoding $r$ dimensional lexical verb vectors into $r \times r$ verb matrices, and present empirical results for each. We use the additional structure that the kronecker product provides to represent the relational nature of transitive verbs. The results are an indication that  the extra information contained in this larger space contributes to higher quality composition.

One way to encode an $r$-dimensional vector as a $r\times r$ matrix is to embed it as the diagonal of that matrix. It remains  open to decide what the non-diagonal values should be. We experimented with 0s and 1s as padding values. If the vector of the verb is $[c_1, c_2, \cdots, c_r]$ then for the 0 case (referred to as {\bf 0-diag}) we obtain the following matrix:
\[
\underline{\text{tverb}} \quad = \quad 
\left (
\begin{array}{ccccc} 
c_1 &0& \cdots & 0\\
0 & c_2 & \cdots & 0\\
\vdots& \vdots & \ddots & \vdots \\
0&0&\ldots&c_r
\end{array}
\right)
\] 
For the 1 case (referred to as {\bf 1-diag}) we obtain the following matrix:
\[
\underline{\text{tverb}} \quad = \quad 
\left (
\begin{array}{ccccc} 
c_1 &1& \cdots & 1\\
1 & c_2 & \cdots & 1\\
\vdots& \vdots & \ddots & \vdots \\
1&1&\ldots&c_r
\end{array}
\right)
\] 
We also considered a third case where the vector is encoded into a matrix by taking the kronecker product of the verb vector with itself:
\[
\underline{\text{tverb}} \quad = \quad \overrightarrow{\text{tverb}} \otimes \overrightarrow{\text{tverb}}
\]
So for $\overrightarrow{\text{tverb}} = [c_1, c_2, \cdots, c_r]$ we obtain the following matrix:
\[
\underline{\text{tverb}} \quad = \quad 
\left (
\begin{array}{ccccc} 
c_1 c_1 &c_1 c_2& \cdots& c_1c_r\\
c_2c_1 & c_2c_2 & \cdots& c_2c_r\\
\vdots&\vdots& \ddots& \vdots \\
c_rc_1&c_rc_2&\cdots&c_rc_r
\end{array}
\right)
\]

\section{Degrees of synonymity for sentences}

The degree of synonymity between meanings of two sentences is computed by measuring their geometric distance. In this work, we used the cosine measure. For two sentences `sub$_1$ tverb$_1$ obj$_1$' and `sub$_2$ tverb$_2$ obj$_2$', this is obtained by  taking the \emph{Frobenius} inner product of
$\overrightarrow{\mbox{sub$_1$ tverb$_1$ obj$_1$}}$ and $\overrightarrow{\mbox{sub$_2$ tverb$_2$ obj$_2$}}$. The use of \emph{Frobenius} product rather than the dot product is because the calculation in Equation (\ref{eq}) produces matrices rather than row vectors. We normalized the outputs by the multiplication of the lengths of their corresponding matrices.

\section{Experiment}

In this section, we describe the experiment used to evaluate and compare these three methods. The experiment is on the dataset developed in~\cite{GrefenetalEMNLP}.

\paragraph{Parameters}
We used the parameters described by \newcite{Lapata} for the noun and verb vectors. All vectors were built from a lemmatised version of the BNC. The noun basis was the 2000 most common context words, basis weights were the probability of context words given the target word divided by the overall probability of the context word. These features were chosen to enable easy comparison of our experimental results with those of Mitchell and Lapata's original experiment, in spite of the fact that there may be more sophisticated lexical distributional models available.

\paragraph{Task} 
This is an extension of \newcite{Lapata}'s disambiguation task from intransitive to transitive sentences. The general idea  behind the transitive case (similar to the intransitive one) is as follows: meanings of ambiguous transitive verbs vary based on their subject-object context. For instance the verb `meet' means `satisfied' in the context `the system met the criterion'  and it means `visit', in the context `the child met the house'. Hence if we build meaning vectors for these sentences compositionally, the degrees of synonymity of the sentences can be used to disambiguate the meanings of the verbs in them.  

Suppose a verb has two meanings $a$ and $b$ and that it has occurred in two sentences.  Then if in both of these sentences it has its meaning $a$, the two sentences will have a high degree of synonymity, whereas if in one sentence the verb has meaning $a$ and in the other meaning $b$, the sentences will have a lower degree of synonymity. For instance `the system met the criterion' and `the system satisfied the criterion'  have a high degree of semantic similarity, and similarly for `the child met the house' and `the child visited the house'. This degree decreases for the pair `the child met the house' and `the child satisfied the house'. 

\paragraph{Dataset} 
The dataset is built using the same guidelines as \newcite{Lapata}, using transitive verbs obtained from CELEX\footnote{\texttt{http://celex.mpi.nl/}} paired with subjects and objects. We first picked 10 transitive verbs from the most frequent verbs of the BNC. For each verb, two different non-overlapping meanings were retrieved, by using the JCN (Jiang Conrath) information content synonymity measure of WordNet to select maximally different synsets. For instance  for `meet' we obtained `visit' and `satisfy'. For each original verb, ten sentences containing that verb with the same role were retrieved from the BNC. Examples of such  sentences are `the system met the criterion' and `the child met the house'. For each such sentence, we generated two other related sentences by substituting their verbs by each of their two synonyms.  For instance, we obtained  `the system satisfied the criterion' and `the system visited the criterion' for the first meaning and  `the child satisfied the house' and `the child visited the house' for the second meaning . This procedure provided us with 200 pairs of sentences. 

The dataset was split into four non-identical sections of 100 entries such that each sentence appears in exactly two sections. Each section was given to a group of evaluators who were asked to assign a similarity score to simple transitive sentence pairs formed from the verb, subject, and object provided in each entry (\emph{e.g.}~ `the system met the criterion' from  `system meet criterion').  The scoring scale for human judgement was $[1,7]$, where 1 was most dissimilar and 7 most identical. 

Separately from the group annotation, each pair in the dataset was given the additional arbitrary classification of HIGH or LOW similarity by the authors.

%\medskip
\paragraph{Evaluation Method} 

To evaluate our methods,  we first applied our formulae to compute the similarity of each phrase pair  on a scale of $[0,1]$ and then compared it with human judgement of the same pair. The comparison was performed by measuring Spearman's $\rho$, a rank correlation coefficient ranging from -1 to 1. This provided us with the degree of correlation between the similarities as computed by our model and as judged by human evaluators.  

Following \newcite{Lapata}, we also computed the mean of HIGH and LOW scores.  However, these scores were solely based on  the authors' personal judgements and as such (and on their own)  do not provide a very reliable measure. Therefore, like \newcite{Lapata}, the models were ultimately judged by Spearman's $\rho$. 

The results  are presented in  table~\ref{results}. The additive and multiplicative rows have, as composition operation, vector addition and component-wise multiplication.
%\footnote{\newcite{Lapata} also consider a combined model which is supervised, i.e.~it has free parameters which are optimized by experimenting with the dataset.  We do not have any similar free parameters, hence it does not directly make sense to compare our model to the combined one. Perhaps it would be interesting to study the consequences of entering such parameters into the  categorical  model.}
The \emph{Baseline} is from a non-compositional approach; it is obtained by comparing the verb vectors of each pair directly and ignoring their subjects and objects. The \emph{UpperBound} is set to be inter-annotator agreement.

\begin{table}[h]
\begin{center}

\begin{tabular}{|lll|c|}
\hline
Model & High & \quad Low & \quad $\rho$\\
\hline
\hline
Baseline & 0.47 & \quad 0.44 & \quad 0.16\\
\hline
\hline
Add & 0.90 & \quad 0.90 & \quad 0.05\\
Multiply & 0.67 & \quad 0.59 & \quad 0.17 \\
\hline
\hline
\textbf{Categorical} &&&\\
\hline
\hline
\textbf{Indirect matrix} & \textbf{0.73} & \quad\textbf{0.72} & \quad \textbf{0.21}\\
\hline

\textbf{0-diag matrix} & \textbf{0.67} & \quad\textbf{0.59} & \quad \textbf{0.17}\\
\hline

\textbf{1-diag matrix} & \textbf{0.86} & \quad\textbf{0.85} & \quad \textbf{0.08}\\
\hline
$v \otimes v$ \textbf{matrix} & \textbf{0.34} & \quad\textbf{0.26} & \quad \textbf{0.28}\\
\hline
\hline
UpperBound & 4.80 & \quad 2.49 & \quad 0.62 \\
\hline
\end{tabular}
\end{center}
\label{results}
\caption{Results of  compositional disambiguation.}
\label{table:results}
\end{table}

The {\bf indirect matrix} performed better than the vectors encoded in diagonal matrices padded with 0 and 1. However, surprisingly, the kronecker product of this vector with itself provided better results than all the above. The results were statistically significant with $p<0.05$.

\section{Analysis of the Results}
Suppose the vector of subject is $[s_1, s_2, \cdots, s_r]$ and the vector of object is $\overrightarrow{\text{obj}} = [o_1, o_2, \cdots, o_r]$, then the matrix of $\overrightarrow{\text{sub}} \otimes \overrightarrow{\text{obj}}$ is:

\vspace{-4mm}
 \[
\left (
\begin{array}{ccccc} 
s_1 o_1 &s_1 o_2& \cdots& s_1o_r\\
s_2o_1 & s_2o_2 & \cdots& s_2o_r\\
\vdots&&&\\
s_ro_1&s_ro_2&\cdots&s_ro_r
\end{array}
\right)
 \]
 
 \vspace{-2mm}
 \noindent
After computing Equation (\ref{eq}) for each generation method of $\underline{\text{tverb}}$,  we obtain the following three matrices for the meaning of a transitive sentence:
\[
\text{\bf 0-diag} \colon   \quad 
\left (
\begin{array}{ccccc} 
c_1 s_1o_1 &0& \cdots& 0\\
0 & c_2s_2o_2 & \cdots& 0\\
\vdots & \vdots & \ddots & \vdots \\
0&0&\cdots&c_rs_ro_r
\end{array}
\right)
\]
This method discards all of the non-diagonal information about the subject and object, for example there is no occurrence of $s_1o_2$, $s_2o_1$, etc. 
\[
\text{\bf 1-diag} \colon   \quad 
\left (
\begin{array}{ccccc} 
c_1 s_1o_1 &s_1o_2& \cdots & s_1 o_r\\
s_2o_1 & c_2s_2o_2  & \cdots & s_2 o_r\\
\vdots& \vdots & \ddots & \vdots \\
s_ro_1&s_ro_2&\cdots&c_rs_ro_r
\end{array}
\right)
\]
This method conserves the information about the subject and object, but only applies the information of the verb to the diagonals: $s_1$ and $o_2$, $s_2$ and $o_1$, etc.~are never related to each other via the verb. 
\begin{align*}
v \otimes v\colon\
 \left (
\begin{array}{ccccc} 
c_1 c_1 s_1o_1 & c_1 c_2 s_1o_2 & \cdots & c_1 c_r s_1 o_r\\
c_2 c_1  s_2o_1 & c_2 c_2 s_2o_2  & \cdots & c_2 c_r s_2 o _r\\
\vdots & \vdots & \ddots & \vdots \\
c_r c_1 s_r o_1 & c_r c_2 s_r o_2 & \cdots & c_r c_r s_ro_r
\end{array}
\right)
\end{align*}

This method not only conserves the information of the subject and object, but also applies to them all of the information encoded in the verb. These data propagate to \emph{Frobenius} products when computing the semantic similarity of sentences and justify the empirical results. 

The unexpectedly good performance of the $v \otimes v$ matrix relative to the more complex indirect method is surprising, and certainly demands further investigation. What is sure is that they each draw upon different aspects of semantic composition to provide better results.  There is certainly room for improvement and empirical optimisation in  both of these relation-matrix construction methods. 

Furthermore, the success of both of these methods relative to the others examined in Table \ref{table:results} shows that it is the extra information provided in the matrix (rather than just the diagonal, representing the lexical vector) that encodes the \emph{relational nature} of transitive verbs, thereby validating in part the requirement suggested in \newcite{Coeckeetal} and \newcite{GrefenetalEMNLP} that relational word vectors live in a space the dimensionality of which be a function of the arity of the relation.

%\section{Related Work}
%\newcite{Baroni} and \newcite{Guevara} experiment with supervised models that are similar to \newcite{Lapata}'s combined model. We would like to discover links with these approaches, but do not currently see a direct way of introducing similar parameters into our compositional formulae. 
%%The formula for computing meaning of a sentence comes directly from a general mathematical setting~\cite{Coeckeetal} which formalises the logical structure of category of finite dimensional vector spaces and unifies it with the logical structure of category of pregroup grammars. 
%%NOTE: Is the above sentence really needed? We express the relaiton to this model several times throughout the paper.
%The approach of \newcite{Daoud} is, mathematically speaking, closer to \newcite{Coeckeetal}, but in the context of  infinite dimension vector spaces. One possible way of extending our setting to infinite cases is briefly discussed in \newcite{Grefenetal}.

\end{document}